\documentclass{article}
\usepackage{log_2024}	

\usepackage{booktabs}						
\usepackage{multirow}						
\usepackage{amsfonts}						
\usepackage{graphicx}						
\usepackage{duckuments}						
\usepackage{xcolor}         
\usepackage{xspace}
\usepackage[utf8]{inputenc} 
\usepackage[T1]{fontenc}    
\usepackage{url}            
\usepackage{amsfonts}       
\usepackage{nicefrac}      
\usepackage{microtype}      
\usepackage{xcolor}         
\usepackage{amsmath}        
\usepackage{caption}
\usepackage{subcaption}
\usepackage{tabu}
\usepackage{multirow}
\usepackage{tikz}
\usepackage{filecontents}
\usepackage[ruled,linesnumbered,noend]{algorithm2e}
\usepackage{array, makecell} %
\usepackage{wrapfig}
\usepackage{enumitem}

\usepackage[numbers,compress,sort]{natbib}	

\title[UTG: Towards a Unified View of Snapshot and Event Based Models for Temporal Graphs]{UTG: Towards a Unified View of Snapshot and \\
Event Based Models for Temporal Graphs}

\newcommand{\etc}{et al.\xspace}

\newcommand{\mat}[1]{\ensuremath{\mathbf{#1}}}

\newtheorem{definition}{Definition}

\newcommand{\name}{UTG\xspace}

\newcommand{\first}[1]{\textbf{\textcolor{red}{#1}}}
\newcommand{\second}[1]{\underline{\textcolor{blue}{#1}}}
\newcommand{\third}[1]{\emph{\textcolor{violet}{#1}}}

\newcommand{\changed}[1]{\textcolor{black}{#1}}
\newcommand{\revised}[1]{\textcolor{black}{#1}}

\author[Huang et al.]{
Shenyang Huang\textsuperscript{1,2,}\thanks{Equal contributions.} 
\quad Farimah Poursafaei\textsuperscript{1,2,}\footnotemark[1] \\
\quad \textbf{Reihaneh Rabbany}\textsuperscript{1,2,5} 
\quad  \textbf{Guillaume Rabusseau}\textsuperscript{1,4,5} 
\quad \textbf{Emanuele Rossi}\textsuperscript{3} \\
\textsuperscript{1}Mila - Quebec AI Institute, 
\textsuperscript{2}School of Computer Science, McGill University\\
\textsuperscript{3}VantAI, 
\textsuperscript{4}DIRO, Université de Montréal,
\textsuperscript{5}CIFAR AI Chair \\
}

\begin{document}

\maketitle

\begin{abstract}

Many real world graphs are inherently dynamic, constantly evolving with node and edge additions. 
These graphs can be represented by temporal graphs, either through a stream of edge events or a sequence of graph snapshots. Until now, the development of machine learning methods for both types has occurred largely in isolation, resulting in limited experimental comparison and theoretical cross-pollination between the two. In this paper, we introduce Unified Temporal Graph (\name), a framework that unifies snapshot-based and event-based machine learning models under a single umbrella, enabling models developed for one representation to be applied effectively to datasets of the other. We also propose a novel \name training procedure to boost the performance of snapshot-based models in the streaming setting. We comprehensively evaluate both snapshot and event-based models across both types of temporal graphs on the temporal link prediction task. Our main findings are threefold: first, when combined with \name training, snapshot-based models can perform competitively with event-based models such as TGN and GraphMixer even on event datasets.
Second, snapshot-based models are at least an order of magnitude faster than most event-based models during inference. Third, while event-based methods such as NAT and DyGFormer outperforms snapshot-based methods on both types of temporal graphs, this is because they leverage joint neighborhood structural features thus emphasizing the potential to incorporate these features into snapshot-based models as well. These findings highlight the importance of comparing model architectures independent of the data format and suggest the potential of combining the efficiency of snapshot-based models with the performance of event-based models in the future. 
\end{abstract}

\section{Introduction}

Recently, Graph Neural Networks (GNNs)\citep{kipf2016semi,velivckovic2018graph} and Graph Transformers\cite{ying2021transformers,rampavsek2022recipe} have achieved remarkable success in various tasks for static graphs, such as link prediction, node classification, and graph classification~\cite{hu2020open}. These successes are driven by standardized empirical comparisons across model architectures~\cite{hu2020open} and theoretical insights into the expressive power of these models~\cite{xu2018powerful}.

However, real-world networks such as financial transaction networks~\cite{shamsi2022chartalist}, social networks~\cite{nadiri2022large}, and user-item interaction networks~\cite{kumar2019predicting} are constantly evolving and rarely static. These evolving networks are often modeled by Temporal Graphs (TGs), where entities are represented by nodes and temporal relations are represented by timestamped edges between nodes. Temporal graphs are categorized into two types: Discrete-Time Dynamic Graphs (DTDGs) and Continuous-Time Dynamic Graphs (CTDGs)~\cite{kazemi2020representation}. DTDGs are represented by an ordered sequence of graph snapshots, while CTDGs consist of timestamped edge streams. Both representations of temporal graphs are prevalent in real-world applications.

Until now, the development of ML methods for both types has occurred mostly independently, resulting in limited experimental comparison and theoretical cross-pollination between the two. We argue that the time granularity of the data collection process together with the requirements of the downstream task have created a gap between DTDG and CTDG in \emph{model development} and \emph{evaluation}.

\textbf{Isolated Model Development.} Despite the similarities between DTDGs and CTDGs, models for these graphs have been developed largely in isolation. Adopting the terminology of~\cite{longa2023graph}, models targeting DTDGs focus on learning from a sequence of graph snapshots (\emph{snapshot-based models})~\cite{htgn2021,gclstm2022,evolvegcn2020} , while methods for CTDGs focus on learning from a stream of timestamped edge events (\emph{event-based models})~\cite{tgn2020,poursafaei2022towards,luo2022neighborhood}. The disparate data representations of DTDGs and CTDGs have impeded comprehensive comparison across models developed for each category. Consequently, there are limited theoretical insights and empirical evaluations of the true potential of these models when compared together. In real-world applications, representing the data as CTDGs or DTDGs is often a design choice, 
and the ambiguity of the actual performance merits of both categories makes it challenging to select the optimal model in a practical setting.

\textbf{Distinct Evaluation Settings.} Another obstacle to comparing snapshot and event-based methods is their distinct evaluation settings. \changed{Snapshot-based methods have been primarily tested under the \emph{deployed setting}~\cite{htgn2021,evolvegcn2020}}~\footnote{An exception to this is ROLAND~\cite{roland2022}, which proposed the live-update setting (see Appendix~\ref{app:eval} for a more detailed discussion).}, where the test set information is strictly not available to the model, and training set information is used for prediction. In contrast, event-based models are designed for the \emph{streaming setting}~\cite{tgn2020,huang2023temporal}, where streaming predictions allow the model to use recently observed information, enabling event-based models to update their node representations at test time.

In this work, we aim to bridge the gap between event-based and snapshot-based models by providing a unified framework to train and evaluate them to predict future events on any type of temporal graph. Our main contributions are as follows:
\begin{itemize}
\item \textbf{Unified framework}: We propose \emph{Unified Temporal Graph} (\name), a framework that unifies snapshot-based and event-based temporal graph models under a single umbrella, enabling models developed for one representation to be applied effectively to datasets of the other.

\item \textbf{Updating snapshot-based models}: We propose a novel \name training strategy to boost the performance of snapshot-based models in the streaming setting. This allows snapshot-based models to achieve competitive performance with event-based models such as TGN and GraphMixer on the \texttt{tgbl-wiki} and Reddit CTDG datasets.

\item \textbf{Benchmarking}: By leveraging the \name framework, we conduct the first systematic comparison between snapshot and event-based models on both CTDG and DTDG datasets. While some event-based methods such as NAT and DyGFormer outperform snapshot-based methods on both CTDGs and DTDGs, we posit this is due to leveraging joint neighborhood structural features rather than a fundamental property of event-based methods. Additionally, snapshot-based methods are at least an order of magnitude faster than event-based methods while achieving competitive performance. This suggests several future directions, such as integrating joint neighborhood structural features in snapshot-based models and developing a universal method that combines accuracy and efficiency for both DTDGs and CTDGs.
\end{itemize}

\textbf{Reproducibility:} The code and data for this project is publicly available on Github: \\
\url{https://github.com/shenyangHuang/UTG}

\section{Related Work}

\changed{Holme \etc~\cite{holme2015modern} provided a general overview of the many types of real world temporal networks showing that temporal graphs are ubiquitous in many applications.}
Recently, many ML methods were developed for temporal graphs.
The well-adopted categorization by \citet{kazemi2020representation} are defined by the two types of temporal graphs: Discrete Time Dynamic Graphs~(DTDGs) and Continuous Time Dynamic Graphs~(CTDGs). \changed{Due to difference in input data format, empirical comparison between methods designed for CTDGs and DTDGs are under-explored and these two categories are often considered distinct lines of research despite many similarities in models' design.}
More detailed discussions on related work can be found in Appendix~\ref{app:method}.


\textbf{Discrete Time Dynamic Graphs.} Early methods often represent temporal graphs as a sequence of graph snapshots while adapting common graph neural networks such as Graph Covolution Network~(GCN)~\cite{kipf2016semi} used in static graphs for DTDGs. 
For example, EGCN~\cite{evolvegcn2020} employs a Recurrent Neural Network~(RNN) to evolve the parameters of a GCN over time. In comparison, GCLSTM~\cite{gclstm2022} learns the graph structure via a GCN while capturing temporal dependencies with an LSTM network~\cite{hochreiter1997long}. Pytorch Geometric-Temporal (PyG Temporal)~\cite{pygtemporal2021} is a comprehensive framework that facilitate neural spatiotemporal signal processing which implements existing work such as EGCN and GCLSTM in an efficient manner. HTGN~\cite{htgn2021} utilizes hyperbolic geometry to better capture the complex and hierarchical nature of the evolving networks. Recently, \citet{roland2022} introduced a novel \emph{live-update setting} where GNNs are always trained on the most recent observed snapshot after making predictions. In comparison, the \emph{streaming setting} in this work allows the model to use observed snapshots for forward pass but no training are permitted on test set. \changed{Zhu \etc~\cite{zhu2023wingnn} designed the WinGNN framework for the live-update setting where a simple GNN with meta-learning strategy is used in combination with a novel random gradient aggregation scheme, removing the need for temporal encoders. }

\textbf{Continuous Time Dynamic Graphs.} Event-based methods process temporal graphs as a stream of timestamped edges. DyRep~\cite{trivedi2019dyrep} and JODIE~\cite{kumar2019predicting} are two pioneering work on CTDGs. 
TGAT~\cite{tgat2020} is one of the first work for studying inductive representation learning on temporal graphs. \citet{tgn2020} introduce Temporal Graph Networks (TGNs), a generic inductive framework of Temporal Graph Networks, showing DyRep, JODIE and TGAT as its special cases. Methods such as CAWN~\cite{cawn2021} and NAT~\cite{luo2022neighborhood} both focuses on learning the joint neighborhood of the two nodes of interest in the link prediction task. CAWN focuses on learning from temporal random walks while NAT is a neighborhood-aware temporal network model that introduces a dictionary-type neighborhood representation for each node. TCL~\cite{tcl2021} and DyGFormer~\cite{dygformer2023} applies transformer based architecture on CTDGs, inspired by the success of transformer based architectures on time series~\cite{wen2023transformers}, images~\cite{dosovitskiy2020image} and natural language processing~\cite{brown2020language}. \changed{Despite the promising performance of event-based methods, recent work showed significant limitations in the standard link prediction evaluation due to the simplicity of negative samples used for evaluation~\cite{poursafaei2022towards}. To improve the evaluation for CTDG, \citet{huang2023temporal} proposed the Temporal Graph Benchmark~(TGB), a collection of large-scale and realistic datasets from distinct domains for both link and node level tasks. }


\section{Preliminaries}
\label{sec:preliminaries}

\begin{definition}[Continuous Time Dynamic Graphs] 
A Continuous Time Dynamic Graph~(CTDG) $\mathcal{G}$ is formulated as a collection of edges represented as tuples with source node, destination node, start time and end time;
\begin{equation*}
\mathcal{G}=\{ (s_0, d_0, t_0^{s}, t_0^{e}), (s_1, d_1, t_1^{s}, t_1^{e}), \dots , (s_k,d_k,t_k^{s}, t_k^{e}) \}
\end{equation*} where, for edge $i\in[0,k]$, $s_i$ and $d_i$ denote source and destination respectively. The start times are ordered chronologically $t_0^s \leq t_1^s \leq ... \leq t_k^s$, each start time is less than or equal to the corresponding end time $t_i^s \leq t_i^e$, hence for each timestamp $t$ we have  $t \in [t_0^s, t_k^e]$. 
\end{definition}%
Without loss of generality, one can normalize the timestamps in $\mathcal{G}$ from $[t_0, t_k]$ to $[0, 1]$ by applying $t = \frac{t - t_0}{t_k} \forall t \in [t_0, t_k]$.
Real world temporal networks can be broadly classified into two inherent types based on the nature of their edges: \revised{\emph{transient networks} and \emph{persistent networks}.} 
Examples of spontaneous networks include transaction networks, retweet networks, Reddit networks, and other activity graphs. 
Here, the edges are spontaneous thus resulting in the start time and end time of an edge being the same, i.e. $t^{s} = t^{e}$. This formulation is inline with related studies in~\cite{poursafaei2022towards,kazemi2020representation,huang2023temporal, scholtes2017network}. For relationship networks such as friendship networks, contact networks, and collaboration networks, the edges often persist over a period of time resulting in $t^{s} \neq t^{e}$.

\begin{definition}[Discrete Time Dynamic Graphs] 
A Discrete Time Dynamic Graph~(DTDG) $\mat{G}$ is a sequence of graph snapshots sampled at regularly-spaced time intervals~\cite{kazemi2020representation}:
\begin{equation*}
\mat{G} = \{ \mat{G}_0, \mat{G}_1, \dots, \mat{G}_T \}
\end{equation*}  $\mat{G}_t = \{\mat{V}_t, \mat{E}_t\}$ is the graph at snapshot $t\in [0,T]$, where $\mat{V}_t$, $\mat{E}_t$ are the set of nodes and edges in $\mat{G}_t$, respectively. 
\end{definition}
\section{\name Framework} \label{sec:method}


\begin{wrapfigure}{R}{0.5\columnwidth}
\centering
    \includegraphics[width=0.5\columnwidth]{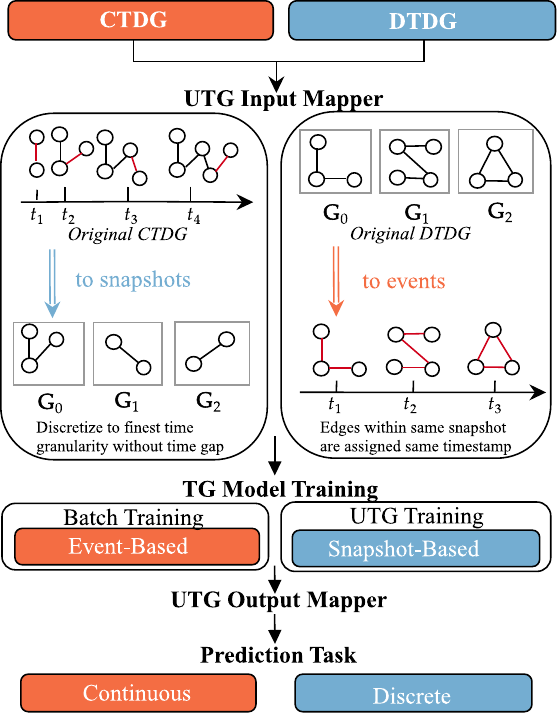}
    \caption{\revised{Illustration of the \name framework. The input graph is processed by the \name input mapper to generate the appropriate input data format for TG models. The model predictions are then processed by the \name output mapper for prediction. }}
\label{Fig:UTG}
\vspace{-30pt}
\end{wrapfigure}

In this section, we present the Unified Temporal Graph~(\name) framework which aims to unify snapshot-based and event-based temporal graph models under the same framework, enabling temporal graph models to be applied to both CTDGs and DTDGs. \name has two key components: \emph{input mapper} and \emph{output mapper}. Input mapping converts the input temporal graph into the appropriate representation needed for a given method, i.e. snapshots or events. Output mapper transforms the prediction of the model to the required time granularity of the task. Figure~\ref{Fig:UTG} shows the workflow of \name framework. \name enables any temporal graph learning methods to be applied to any input temporal graph. 


\subsection{\name Input Mapper} \label{sub:input_map}

Both snapshot and event-based TG methods require specific input data format. For snapshot-based models, discretizing CTDG data into a sequence of snapshots is required. For event-based models, DTDG snapshots need to be converted into batches of events. 


\paragraph{Converting CTDG to Snapshots} Here, we formulate the discretization process which converts a continuous-time dynamic graph into a sequence of graph snapshots for snapshot-based models.

\begin{definition}[Discretization Partition] 
Let $0$ and $1$ be the normalized start and end time of a temporal graph $\mathcal{G}$.
A discretization partition $\mathbf{P}$ of the interval $[0, 1]$ is a collection of intervals: 
\begin{equation*}
    \mathbf{P} = \{ [\tau_0, \tau_1], [\tau_1, \tau_2], \dots, [\tau_{k-1}, \tau_k] \}
\end{equation*} such that $0 = \tau_0 < \tau_1 < \dots < \tau_k = 1$ and where $k \in \mathbb{N}$.
\end{definition}

Hence, a discretization partition $\mathbf{P}$ defines a finite collection of non-overlapping intervals and its norm is defined as:
\begin{equation*}
    || \mathbf{P} || = \max \{ |\tau_1 - \tau_0|, |\tau_2 - \tau_1|, \dots, |\tau_k - \tau_{k-1}| \}
\end{equation*}
The norm $|| \mathbf{P} ||$ can also be interpreted as the \emph{max duration} of a snapshot in the temporal graph $\mathcal{G}$. The cardinality of $\mathbf{P}$ is denoted by $|\mathbf{P}|$.

\begin{definition}[Regular Discretization Partition] 
A given discretization partition $\mathbf{P}$ is regular if and only if:
\begin{equation*} 
    \forall [\tau_i, \tau_j] \in \mathbf{P}, \;|\tau_j - \tau_i| = || \mathbf{P} || = \frac{|\tau_k - \tau_0|}{|\mathbf{P}|}
\end{equation*} 
\end{definition}%
In this case, all intervals have the same duration equal to  $||\mathbf{P}||$.

\begin{definition}[Induced Graph Snapshots]
Given a Continuous Time Dynamic Graph $\mathcal{G}$ and a Regular Discretization Partition $\mathbf{P}$, the Induced Graph Snapshots $\mat{G}$ are formulated as:
\begin{equation*}
    \mat{G} = \{ \mat{G}_{\tau_0}^{\tau_1}, \mat{G}_{\tau_1}^{\tau_2}, \dots, \mat{G}_{\tau_{k-1}}^{\tau_k} \}
\end{equation*} 
where $\mat{G}_{\tau_i}^{\tau_j}$ is defined as the aggregated graph snapshot containing all edges that have \revised{a \emph{start time} $t_s < \tau_j$} and an \emph{end time} $t_e \geq \tau_i$, i.e. edges that are present solely within the \revised{$[{\tau_i},{\tau_j})$} interval.  
\end{definition}
Note that for spontaneous networks, each edge exist at a specific time point $t_s = t_e$ thus only belonging to a single interval/snapshot. For relationship networks however, it is possible for an edge to belong to multiple intervals depending on its duration.

\begin{definition}[Discretization Level]
Given a regular discretization partition $\mathbf{P}$ and the timestamps in a temporal graph $\mathcal{G}$ normalized to $[0,1]$, the \emph{discretization level} $\Delta$ of $\mathbf{P}$ is computed as :
\begin{equation*}
    \Delta = \frac{1}{|\mathbf{P}|}
\end{equation*} 
where $|\mathbf{P}|$ is the cardinality of the partition or the number of intervals. \end{definition}
Note that $\Delta \in [0,1]$. When $|\mathbf{P}|=1$ then $\Delta = 1$ which means the temporal graph is collapsed into a single graph snapshot (i.e. a static graph). On the other extreme, we have $ \lim_{|\mathbf{P}|\to\infty} \Delta = 0$, preserving the continuous nature of the continuous time dynamic graph $\mathcal{G}$. 

\begin{definition}[Time Gap]
Given a Continuous Time Dynamic Graph $\mathcal{G}$ and a Regular Discretization Partition $\mathbf{P}$, a Time Gap occurs when there exist one or more snapshots in the Induced Graph Snapshots $\mat{G}$ with an empty edge set.
\end{definition}
In this work, we choose the number of intervals in discretization by selecting the finest time granularity which would not induce a time gap. This ensures that there are no empty snapshots in the induced graph snapshots.


\paragraph{Converting DTDG to Events}

While it may seem straightforward to convert DTDG to events --- simply create one event with timestep $t$ for each edge in snapshot $G_t$ --- some subtleties related to batch training and memory update of event-based models have to be considered to avoid data leakage. 
Event-based models often receive batches of events (or edges) with a fixed dimension as inputs~\cite{tgn2020,dygformer2023,luo2022neighborhood}. In discrete-time dynamic graphs, all edges in a snapshot have the same timestamp and are assumed to arrive simultaneously. Therefore, using a fixed batch size can result in splitting the snapshot into multiple batches. Because models in the \emph{streaming setting}~\cite{huang2023temporal} such as TGN~\cite{tgn2020} and NAT~\cite{luo2022neighborhood} update their representation of the temporal graph at the end of each batch, predicting a snapshot across multiple batches leads to data leakage: a portion of the edges from the snapshot is used to predict other (simultaneous) edges from the same snapshot. To avoid data leakage on DTDGs, we ensure that each snapshot is contained in a single batch for event-based models\footnote{In case the resulting batch would not fit in memory, one can delay the memory update (and parameter updates during training) only after all edges from the current snapshot have been processed, something akin to gradient accumulation.}. \revised{Note that the issue of splitting edges that share the same timestamp into multiple batches can exist in general CTDG datasets as well, more likely for datasets with a large burst of edges at a single timestamp.}

\subsection{\name Output Mapper}

The output task on the temporal graph can be either discrete or continuous. Discrete tasks refer to predicting which edges will be present at a future snapshot~(with an integer timestep). Continuous tasks refer to predicting which edges will be present for a given UNIX timestamp in the future. 
Snapshot-based models often omits the timestamp of the prediction as an input, implicitly assuming the prediction is for the next snapshot. Therefore, applying snapshot-based models for a continuous task requires 1). always updating the model with all the information available until the most recent observed snapshot (\textit{test-time update}) and 2). mapping the discrete-time prediction to a continuous timestamp. We explain here how to map the prediction to a continuous space with zero-order hold. 

\begin{definition}[Zero-order Hold]
    A discrete time signal $y[i]$, $i\in\mathbb N$, can be converted to 
    a continuous time signal $y(t)$, $t\in\mathbb R$,
    by broadcasting the value y[i] as a constant in the interval $[\tau_i,\tau_j]$:
\begin{equation*}
    y(t) = y[i], \ \ \text{for all} \quad \tau_i \leq t \leq \tau_j
\end{equation*}
where $[\tau_i,\tau_j]$ specifies the duration of the discrete signal.
\end{definition}
By applying zero-order hold for snapshot-based models, the predictions can now be broadcasted for a period of time (specifically for the duration of a given snapshot $[\tau_i, \tau_j]$). Therefore, it is now possible to utilize snapshot-based models on continuous-time dynamic graphs. \revised{Note that often snapshot-based model are designed to predict for the immediate next snapshot and not capable of predicting snapshots in more distant future. Therefore, inherently their ability to predict events in the far future is limited when compared to event-based model that explicitly takes a timestamp as input. With zero-order hold, we assume that the event to predict next is within the next snapshot to circumvallate the aforementioned limitation of snapshot-based model. To achieve this, we select the finest time granularity on the CTDG datasets which results in no time gap to construct snapshots. }


\subsection{\revised{Streaming Evaluation}}

\begin{wrapfigure}{R}{0.6\columnwidth}
\vspace{-30pt}
\centering
    \includegraphics[width=0.6\columnwidth]{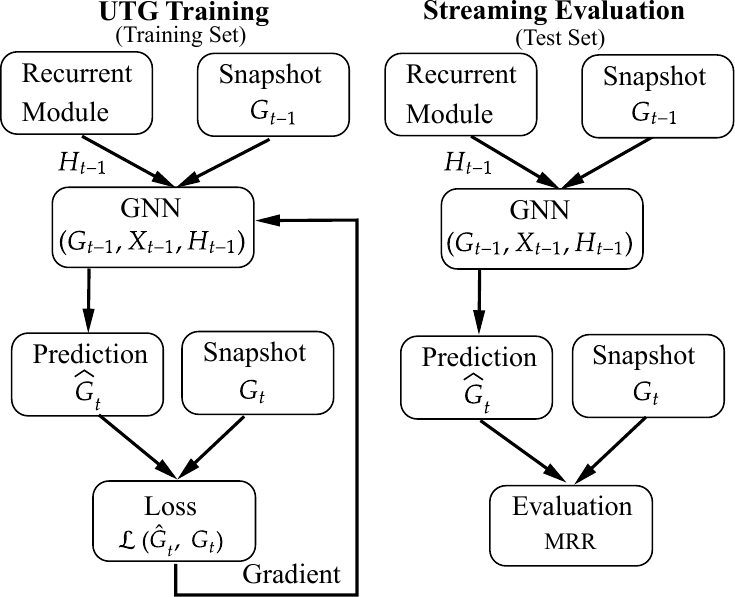}
    \caption{\revised{\name training and evaluation workflow. \name training enables snapshot-based models to perform better for the streaming setting.}}
\label{Fig:train_eval}
\vspace{-15pt}
\end{wrapfigure}


Event-based models often evaluate with the \emph{streaming setting}~\cite{tgn2020,dygformer2023,cawn2021,luo2022neighborhood}. In this setting, information from the previously observed batches of events (or graph snapshot) can be used to update the model however no information from the test set is used to train the model. \revised{In comparison, snapshot-based models are often evaluated in either the \emph{live-update setting}~\cite{roland2022} or the \emph{deployed setting}~\cite{evolvegcn2020,htgn2021}. Detailed differences between evaluation settings are discussed in Appendix~\ref{app:eval}.}

\revised{
In this work, to provide a unified comparison, we focus on the widely used streaming setting for experimental evaluation as it closely resembles real-world settings where after the model is trained, it is required to incorporate newly observed information into its predictions. Note that the UTG framework can also be applied to test under the deployed setting with little changes. For the live-update setting, the changes are required for each model's training procedure as the training set is not split chronologically but rather being a subgraph of each observed snapshot.} Figure~\ref{Fig:train_eval}
shows the evaluation pipeline used for snapshot-based models in \name. After a snapshot is observed, it can be used to update the node representation of the snapshot-based models for the prediction of the next snapshot (only forward pass for inference).

\subsection{\revised{\name Training for Snapshot-based Models}} \label{sec:improve}

Here, we discuss the changes to the snapshot-based methods in the \name framework. \revised{Figure~\ref{Fig:train_eval} illustrates the work flow of \name training for snapshot-based models. The standard training for snapshot-based models, e.g. with Pytorch Geometric Temporal~\cite{rozemberczki2021pytorch}, are designed for tasks such as graph regression or node classification, for the deployed setting. Therefore, the training procedure needs to be adapted accordingly for the link prediction with the streaming setting.}

\revised{
\textbf{Difference of \name Training.}
The first required change is that in standard training, the snapshots up until time $t$ is used as input for predictions at time $t$. This is feasible for node classification and graph regression tasks: the graph structure at time $t$ is used to predict the unknown target labels. However, this is problematic for the link prediction task as the target itself (the graph structure) is not available as input to the model. Therefore, to account for this, we modified the training to use only snapshots up to $\mat{G}_{t-1}$ as input to predict the graph structure at the current step $\mat{G}_{t}$.}

\revised{
The second change is that, in \name, the loss is backpropagated to the model at each snapshot. This is in contrast with accumulating the loss through the whole sequence and only backpropagation once at the end of training (as seen in standard training). This change is motivated by the training procedure often seen in event-based models for the streaming setting. For example, in TGN~\cite{tgn2020}, loss on each batch is computed based on information from the previous batch. \name training enhances the performance of snapshot-based models for the streaming setting and in Section~\ref{sec:experiments}, we demonstrate the performance advantage of \name training on a number of snapshot-based models.}

\revised{
\textbf{Connection to Truncated Backprop Through Time.} By considering the temporal graph as a sequence (of snapshots), the link prediction problem can be seen as a time series prediction task where the information until time $(t-1)$ is used to predict for time $t$. Many snapshot-based models utilize a RNN to model temporal dependency. In this view, accumulating gradients throughout the whole sequence before backpropagation is equivalent to the classical backpropagation through time algorithm~\cite{werbos1990backpropagation}. From this perspective, backpropagating the loss at each snapshot in UTG training can be interpreted as using the Truncated Backpropagation Through Time~(TBTT) algorithm to train the model, with a window size of one. It is known that TBTT helps circumvent common issues of training RNNs such as exploding memory usage and vanishing gradient problem~\cite{bengio1993problem,pascanu2013difficulty}. Therefore, \name training might help alleviate the vanishing gradient problem.}

\section{Experiments} \label{sec:experiments}

In this section, we benchmark both snapshot-based and event-based methods across both CTDG and DTDG datasets under the \name framework. 

\textbf{Datasets.} In this work, we consider five discrete-time dynamic graph datasets and three continuous-time dynamic graph datasets. \texttt{tgbl-wiki} and \texttt{tgbl-review} are datasets from TGB~\cite{huang2023temporal} while the rest are found in \citet{poursafaei2022towards}. The dataset statistics are shown in Table~\ref{tab:data}. The time granularity or discretization level of each DTDG dataset is selected as the finest time granularities where there are no time gaps. The surprise index is defined as $surprise = \frac{|E_{\text{test}} \setminus E_{\text{train}}|}{E_{\text{test}}}$~\cite{poursafaei2022towards} which measures the proportion of unseen edges in the test set when compared to the training set. 


\begin{table*}[t]
\caption{Dataset statistics. } \label{tab:data} 
  \resizebox{\linewidth}{!}{%
  \begin{tabular}{l | l | r r r l l l l l r }
  \toprule
  & Dataset &  \# Nodes & \# Edges
  & \# Unique Edges & Surprise
  & Time Granularity & \# Snapshots \\ 
  \midrule

  \multirow{5}{*}{\rotatebox[origin=c]{90}{\small{DTDG}}} 
  & UCI & 1,899 & 26,628 & 20,296 & 0.535  & Weekly  & 29 \\
  & Enron & 184 & 10,472 & 3,125 & 0.253 & Monthly & 45 \\
  & Contact  & 694 & 463,558 & 79,531 & 0.098 & Hourly & 673\\
  & Social Evo. & 74 & 87,479 & 4,486 & 0.005 & Daily & 244\\ 
  & MOOC & 7,144 & 236,808 & 178,443 & 0.718  & Daily & 31\\  
  \midrule
  \multirow{3}{*}{\rotatebox[origin=c]{90}{\small{CTDG}}} & \texttt{tgbl-wiki} & 9,227 & 157,474 & 18,257 & 0.108  &  UNIX timestamp  & 745 (Hourly)\\ 
  & \texttt{tgbl-review} & 352,637 & 4,873,540 & 4,730,223 & 0.987  &  UNIX timestamp  & 237 (Monthly) \\ 
  & Reddit & 10,984 & 672,447 & 78,516 & 0.069 & UNIX timestamp & 745 (Hourly) \\  
  \bottomrule
  \end{tabular}
  }
\vskip -0.2in
\end{table*}


\changed{\textbf{Evaluation Setting.} A common approach for evaluating dynamic link prediction tasks is similar to binary classification, where one negative edge is randomly sampled for each positive edge in the test set, and performance is measured using metrics like the Area Under the Receiver Operating Characteristic curve (AUROC) or Average Precision (AP)~\cite{tgn2020, tgat2020}. However, recent studies have shown that such evaluation is overly simplistic and tends to inflate performance metrics for most models~\cite{poursafaei2022towards, huang2023temporal}. One main reason is that randomly sampled negative edges are too easy and the more challenging \emph{historical negatives} (past edges absent in the current timestamp) are rarely sampled~\cite{poursafaei2022towards}.
To address these issues, several improvements have been proposed, framing the problem as a ranking task where the model must identify the most probable edge from a large pool of negative samples as well as adding challenging negatives. Therefore, we adopt the improved evaluation methodology used in TGB~\cite{huang2023temporal}, where link prediction is treated as a ranking problem and the Mean Reciprocal Rank (MRR) metric is applied. This metric calculates the reciprocal rank of the true destination node among a large number of possible destinations.}

\changed{
For each dataset, we generate a fixed set of negative samples for each positive edge consisting of 50\% \emph{historical negatives} and 50\% \emph{random negative}, same as in~\cite{huang2023temporal}. For DTDG datasets, we generate 1000 negative samples per positive edge. For TGB datasets, we use the same set of negatives provided in TGB and for Reddit, we generate 1000 negatives similar to before. For graphs with less than 1k nodes, we generate negative samples equal to number of nodes.  We follow the \emph{streaming setting} where the models are allowed to update their representation at test time while gradient updates are not permitted. For DTDG datasets, we select the best results from learning rate $0.001$ or $0.0002$. For CTDG datasets, we report the results from TGB~\cite{huang2023temporal} where available or by learning rate $0.0002$.}

\textbf{Compared Methods.} We compare four event-based methods including TGN~\cite{tgn2020}, DyGFormer~\cite{dygformer2023}, NAT~\cite{luo2022neighborhood} and GraphMixer~\cite{cong2022we}. We also include Edgebank~\cite{poursafaei2022towards}, a scalable and non-parameteric heuristics. In addition, we compare three existing snapshot-based methods including HTGN~\cite{htgn2021}, GCLSTM~\cite{gclstm2022}, EGCNo~\cite{evolvegcn2020}, \revised{and ROLAND-GRU \cite{roland2022}}. Lastly, we adapt a common 2-layer (static) GCN~\cite{kipf2016semi} under the \name framework to demonstrate the flexibility of \name (without a recurrent module). If a method runs out of memory on a NVIDIA A100 GPU~(40GB memory), it is reported as out of memory~(OOM). If a method runs for more than 5 days, it is reported as out of time (OOT).

With the \name framework, we can now compare snapshot-based methods and event-based methods on any temporal graph dataset. This comparison allows us to focus on analyzing the strength and weaknesses of the model design, independent of the data format. 

\begin{table*}[t]
\caption{\revised{Test MRR comparison for snapshot and event-based methods on DTDG datasets, results reported from 5 runs.} Top three models are marked by \first{First}, \second{Second}, \third{Third}.}
\label{tab:dtdg_results}
\centering
\resizebox{\linewidth}{!}{%
\begin{tabu}{l | l | l l l l l}
\toprule
 & Method & UCI & Enron & Contacts  & Social Evo. & MOOC \\
 \midrule
 \multirow{6}{*}{\rotatebox[origin=c]{90}{\textbf{event}}} & 
 TGN~\cite{tgn2020}                                   & 0.091 \scriptsize{$\pm$ 0.002}               & 0.191 \scriptsize{$\pm$ 0.027}           & 0.153 \scriptsize{$\pm$ 0.007}            & 0.283 \scriptsize{$\pm$ 0.009}            & \second{0.174} \scriptsize{$\pm$ 0.009} \\
 & DyGFormer~\cite{dygformer2023}                                 & \second{0.334} \scriptsize{$\pm$ 0.024}   & \first{0.331} \scriptsize{$\pm$ 0.010}    & \first{0.283} \scriptsize{$\pm$ 0.006}    & \first{0.366} \scriptsize{$\pm$ 0.004}    & OOM                             \\
 & NAT~\cite{luo2022neighborhood}                                 & \first{0.356} \scriptsize{$\pm$ 0.048}    & \third{0.276} \scriptsize{$\pm$ 0.014}    & \second{0.245} \scriptsize{$\pm$ 0.015}    & 0.258 \scriptsize{$\pm$ 0.036}            & \first{0.283} \scriptsize{$\pm$ 0.058} \\
 & GraphMixer~\cite{cong2022we}                                   & 0.105 \scriptsize{$\pm$ 0.008}            & \second{0.296} \scriptsize{$\pm$ 0.019}   & 0.055 \scriptsize{$\pm$ 0.003}    & 0.157 \scriptsize{$\pm$ 0.005}            & OOM                             \\
& $\text{EdgeBank}_{\infty}$~\cite{poursafaei2022towards}          & 0.055 & 0.115 & 0.016 & 0.049 & 0.040 \\
& $\text{EdgeBank}_{\text{tw}}$~\cite{poursafaei2022towards}     & \third{0.165} & 0.157 & 0.050 & 0.070 & 0.070 \\
 \midrule
\multirow{4}{*}{\rotatebox[origin=c]{90}{\textbf{snapshot}}} &
HTGN (\name)~\cite{htgn2021}                              & 0.093 \scriptsize{$\pm$ 0.012}        & 0.267 \scriptsize{$\pm$ 0.007}               & 0.165 \scriptsize{$\pm$ 0.001}            & 0.228 \scriptsize{$\pm$ 0.003}            & 0.093 \scriptsize{$\pm$ 0.005} \\  
& GCLSTM (\name)~\cite{gclstm2022}                                  & 0.093 \scriptsize{$\pm$ 0.006}        & 0.170 \scriptsize{$\pm$ 0.008}             & 0.128 \scriptsize{$\pm$ 0.004}            & \third{0.286} \scriptsize{$\pm$ 0.003}    & \third{0.143} \scriptsize{$\pm$ 0.006}  \\
& EGCNo (\name)~\cite{evolvegcn2020}                          & 0.121 \scriptsize{$\pm$ 0.010}        & 0.233 \scriptsize{$\pm$ 0.008}             & \third{0.192} \scriptsize{$\pm$ 0.001}            & 0.253 \scriptsize{$\pm$ 0.006}            & 0.126 \scriptsize{$\pm$ 0.009}  \\
& GCN (\name)~\cite{kipf2016semi}                                   & 0.068 \scriptsize{$\pm$ 0.009}        & 0.164 \scriptsize{$\pm$ 0.011}             & 0.104 \scriptsize{$\pm$ 0.002}            & \second{0.289} \scriptsize{$\pm$ 0.008}   & 0.084 \scriptsize{$\pm$ 0.010}  \\

& \revised{ROLAND (\name)~\cite{roland2022}} & 0.103 \scriptsize{$\pm$ 0.011}        & 0.243 \scriptsize{$\pm$ 0.017}           & 0.145 \scriptsize{$\pm$ 0.002}        & 0.240 \scriptsize{$\pm$ 0.005}   & 0.121 \scriptsize{$\pm$ 0.003}  \\

\bottomrule
\end{tabu}
}
\vskip -0.2in
\end{table*}

\textbf{DTDG Results.}
Table~\ref{tab:dtdg_results} shows the performance of all methods on the DTDG datasets. Surprisingly, we find that event-based methods achieve state-of-the-art performance on the DTDG datasets, particularly with DyGFormer and NAT consistently outperforming other methods. With the improvements from \name, snapshot-based models can obtain competitive performance on datasets such as Enron and Social Evo. Interestingly, even the simple GCN with \name training can achieve second place performance on the Social Evo. dataset. Note that this dataset has the lowest surprise out of all datasets meaning the majority of test set edges have been observed during training, possibly explaining the strong performance of GCN in this case. Lastly, on the MOOC dataset which has the largest number of nodes out of all DTDG datasets, both DyGformer and GraphMixer ran out of memory (OOM) showing their difficulty in scaling with the number of nodes in a snapshot.

\begin{wraptable}{r}{0.6\textwidth}  
  \vspace{-0.18in}
    \centering
 \caption{\revised{Test MRR comparison for snapshot and event-based methods on CTDG datasets, results reported from 5 runs.}  Top three models are marked by \first{First}, \second{Second}, \third{Third}.
 }
  \label{tab:ctdg_results}
  \centering
\resizebox{\linewidth}{!}{%
\begin{tabu}{l | l | l l l}
\toprule

 & Method                                                     & \texttt{tgbl-wiki}  & \texttt{tgbl-review}  & Reddit \\
  \midrule
  \multirow{6}{*}{\rotatebox[origin=c]{90}{\textbf{event}}} & 
 TGN~\cite{tgn2020}                               & 0.396 \scriptsize{$\pm$ 0.060}                & \second{0.349} \scriptsize{$\pm$ 0.020}                 & 0.499 \scriptsize{$\pm$ 0.011}  \\
 & DyGFormer~\cite{dygformer2023}                             & \first{0.798} \scriptsize{$\pm$ 0.004}        & 0.224 \scriptsize{$\pm$ 0.015}                        & OOT                            \\
 & NAT~\cite{luo2022neighborhood}                             & \second{0.749} \scriptsize{$\pm$ 0.010}       & \third{0.341} \scriptsize{$\pm$ 0.020}                & \first{0.693} \scriptsize{$\pm$ 0.015}   \\
 & GraphMixer~\cite{cong2022we}                               & 0.118 \scriptsize{$\pm$ 0.002}                & \first{0.521} \scriptsize{$\pm$ 0.015}                & 0.136 \scriptsize{$\pm$ 0.078}  \\
 & $\text{EdgeBank}_{\infty}$~\cite{poursafaei2022towards}    & 0.495                                         & 0.025                                                 & 0.485                              \\
 & $\text{EdgeBank}_{\text{tw}}$~\cite{poursafaei2022towards} & \third{0.571}                                 & 0.023                                                 & \second{0.589}                              \\
 \midrule
\multirow{4}{*}{\rotatebox[origin=c]{90}{\textbf{snapshot}}} &
 HTGN (\name)~\cite{htgn2021}                       & 0.464 \scriptsize{$\pm$ 0.005}                & 0.104 \scriptsize{$\pm$ 0.002}                          & \third{0.533} \scriptsize{$\pm$ 0.007}    \\   
 & GCLSTM (\name)~\cite{gclstm2022}                           & 0.374 \scriptsize{$\pm$ 0.010}                & 0.095 \scriptsize{$\pm$ 0.002}                        & 0.467 \scriptsize{$\pm$ 0.004}    \\ 
 & EGCNo (\name)~\cite{evolvegcn2020}                   & 0.398 \scriptsize{$\pm$ 0.007}                & 0.195 \scriptsize{$\pm$ 0.001}                        & 0.321 \scriptsize{$\pm$ 0.009}    \\ 
 & GCN (\name)~\cite{kipf2016semi}                            & 0.336 \scriptsize{$\pm$ 0.009}         & 0.186 \scriptsize{$\pm$ 0.002}                        & 0.242 \scriptsize{$\pm$ 0.005}    \\ 

 & \revised{ROLAND (\name)~\cite{roland2022}}   & 0.289 \scriptsize{$\pm$ 0.003}         & 0.297 \scriptsize{$\pm$ 0.006}        & 0.211 \scriptsize{$\pm$ 0.006}    \\ 
\bottomrule
\end{tabu}
}
\vspace{-0.2in}
\end{wraptable}

\textbf{CTDG Results.} 
Table~\ref{tab:ctdg_results} shows the performance of all methods on the CTDG datasets. Similar to DTDG datasets, DyGformer and NAT retains competitive performance here. On the \texttt{tgbl-wiki} and Reddit dataset, HTGN, a snapshot-based model is able to outperform widely-used TGN and GraphMixer model. This shows that learning from the discretized snapshots can be effective even on CTDG datasets. However, snapshot-based models have lower performance on the \texttt{tgbl-review} dataset where the surprise index is high. This shows that the inductive reasoning capability on snapshot-based models should be further improved to generalize to unseen edges. 


\begin{figure}[h]
\centering
\begin{subfigure}{.5\textwidth}
  \centering
  \includegraphics[width=0.8\linewidth]{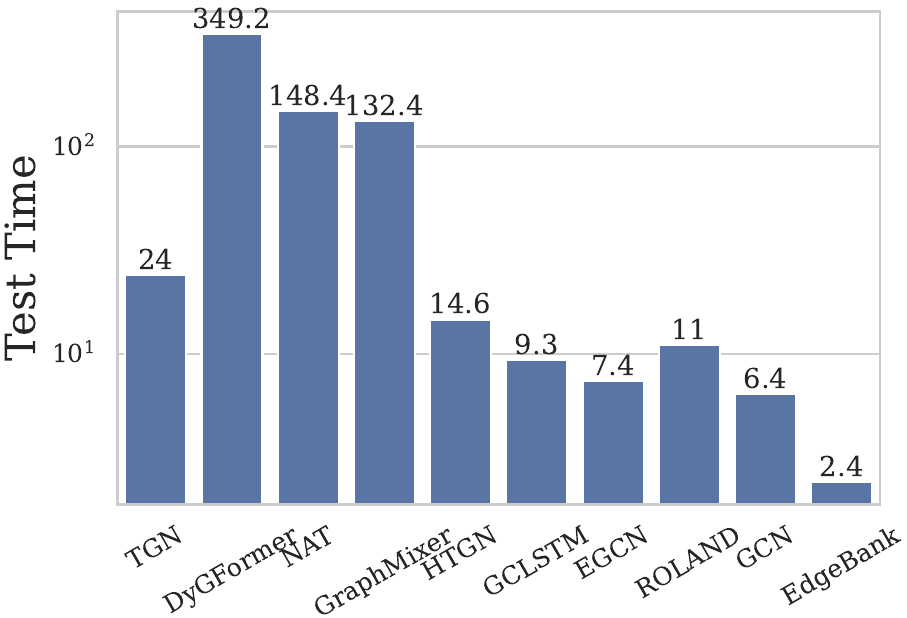}
  \caption{Test time~(seconds) for Social Evo. dataset.}
  \label{Fig:social_evo_time}
\end{subfigure}%
\begin{subfigure}{.5\textwidth}
  \centering
  \includegraphics[width=0.8\linewidth]{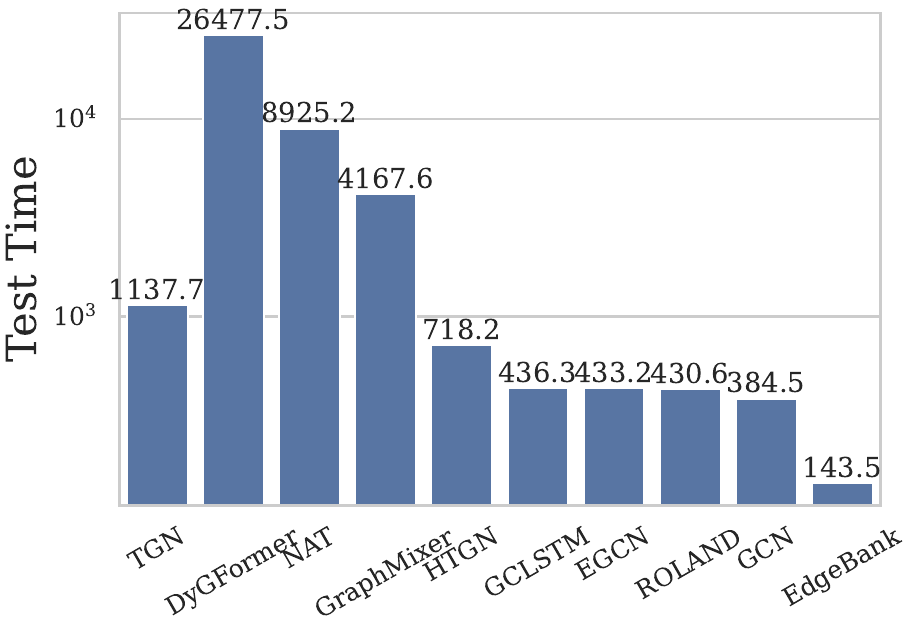}
  \caption{Test time~(seconds) for \texttt{tgbl-review} dataset.} 
  \label{Fig:review_time}
\end{subfigure}
\caption{Snapshot-based models \changed{with \name training} are at least an order of magnitude faster than event-based models for inference. 
}
\vskip -0.1in
\end{figure}

\textbf{Computational Time Comparison}
Figure~\ref{Fig:social_evo_time} and Figure~\ref{Fig:review_time} shows the test inference time comparison for all methods on the Social Evo. and the \texttt{tgbl-review} dataset respectively. The test time for each dataset is reported in Appendix~\ref{app:time}. Overall, we observe that snapshot-based methods are at least an order of magnitude faster than most event-based methods. In comparison, high performing model such as DyGformer has significantly higher computational time thus limiting its scalability to large datasets. One promising direction is to combine the performance of event-based models such as NAT and DyGFormer with that of the efficiency of the snapshot-based models for improved scalability. 


\begin{table*}[t]
\centering
\caption{\revised{Base models with \name training when compared to original training, results averaged across five runs, best results for each model are \textbf{bolded}.}} \label{tab:ablation}
\resizebox{\linewidth}{!}{%
\begin{tabular}{l | l l | l l | l l }
\toprule
 Dataset & \makecell{GCLSTM \\ \footnotesize{(\name)}}
 & \makecell{GCLSTM \\ \footnotesize{(original)}} 
 & \makecell{EGCNo \\ \footnotesize{(\name)}}
 & \revised{\makecell{EGCNo \\ \footnotesize{(original)}}}
 & \makecell{HTGN \\ \footnotesize{(\name)}} 
 & \revised{\makecell{HTGN \\ \footnotesize{(original)}}} \\
 \midrule
UCI         & \textbf{0.093} \scriptsize{$\pm$ 0.006} & 0.047 \scriptsize{$\pm$ 0.006} & \textbf{0.121} \scriptsize{$\pm$ 0.010} & \textbf{0.124} \scriptsize{$\pm$ 0.007} & \textbf{0.093} \scriptsize{$\pm$ 0.012} & 0.052 \scriptsize{$\pm$ 0.006}\\  
Enron       & \textbf{0.170} \scriptsize{$\pm$ 0.006} & 0.131 \scriptsize{$\pm$ 0.003} & \textbf{0.233} \scriptsize{$\pm$ 0.008} &  0.227 \scriptsize{$\pm$ 0.012} & \textbf{0.267} \scriptsize{$\pm$ 0.007} & 0.196 \scriptsize{$\pm$ 0.025} \\ 
Contacts    & \textbf{0.128} \scriptsize{$\pm$ 0.004} & 0.101 \scriptsize{$\pm$ 0.031} & 0.192 \scriptsize{$\pm$ 0.001}  & \textbf{0.195} \scriptsize{$\pm$ 0.001} & \textbf{0.165} \scriptsize{$\pm$ 0.001} & 0.129 \scriptsize{$\pm$ 0.020} \\ 
Social Evo. & \textbf{0.286} \scriptsize{$\pm$ 0.003} & \textbf{0.287} \scriptsize{$\pm$ 0.009} & \textbf{0.253} \scriptsize{$\pm$ 0.006} &  0.231 \scriptsize{$\pm$ 0.009} & \textbf{0.228} \scriptsize{$\pm$ 0.003} & 0.178 \scriptsize{$\pm$ 0.024}\\ 
MOOC        & \textbf{0.143} \scriptsize{$\pm$ 0.006} & 0.076 \scriptsize{$\pm$ 0.003} & \textbf{0.126} \scriptsize{$\pm$ 0.009} & 0.119 \scriptsize{$\pm$ 0.009} & \textbf{0.093} \scriptsize{$\pm$ 0.005} & 0.079 \scriptsize{$\pm$ 0.011} \\ 
\bottomrule
\end{tabular}
}
\vskip -0.2in
\end{table*}

\textbf{Benefits of \name Training.}
\name training enables snapshot-based models to effectively incorporate novel information during inference. \revised{Table~\ref{tab:ablation} shows the performance benefit of using \name training on the GCLSTM, HTGN and EGCNo models when compared to the original training scheme (such as in Pytorch Geometric Temporal) for DTDG datasets. \name training significantly enhances model performance across most datasets and performs identically on the rest.}

\textbf{Discussion.} Event-based methods such as NAT and DyGFormer tend to perform best on both CTDG and DTDG, potentially leading to the premature conclusion that event-based modeling is the preferred paradigm and that snapshot-based models should be avoided. However, the superior performance of NAT and DyGFormer could be primarily due to their ability to leverage joint neighborhood structural features, specifically the common neighbors between the source and destination nodes of a link~\cite{luo2022neighborhood,dygformer2023}. This approach has been shown to be fundamental for achieving competitive link prediction on static graphs~\cite{zhang2018link, zhang2020labeling}.
In contrast, none of the existing snapshot-based methods incorporate joint neighborhood structural features. Therefore, the performance difference could be mainly attributed to this factor rather than an intrinsic difference between event-based and snapshot-based models. This is \revised{suggested} by the fact that event-based models such as TGN and GraphMixer, which omits these features, have no clear performance advantage over snapshot-based methods. Moreover, snapshot-based methods are more computationally efficient and might be preferred when efficiency is important. These considerations suggest that both event-based and snapshot-based methods have their own merit. We believe that combining the strengths of both approaches is an important future direction.
\section{Conclusion}
\label{sec:conclusion}

In this work, we introduce the \name framework, unifying both snapshot-based and event-based temporal graph models under a single umbrella. With the \name input mapper and \name output mapper, temporal graph models developed for one representation can be applied effectively to datasets of the other. To compare both types of methods in the streaming setting for evaluation, we propose the \name training to boost the performance of snapshot-based models. Extensive experiments on five DTDG datasets and three CTDG datasets are conducted to comprehensively compare snapshot and event-based methods. We find that top performing models on both types of datasets leverage joint neighborhood structural features such as the number of common neighbors between the source and destination node of a link. In addition, snapshot-based models can achieve competitive performance to event-based model such as TGN and GraphMixer while being an order of magnitude faster in inference time. Thus, an important future direction is to combine the strength of both types of methods to achieve high performing and scalable temporal graph learning methods.

\section*{Acknowledgements}



This research was supported by the Canadian Institute for Advanced Research (CIFAR AI chair program), Natural Sciences and Engineering Research Council of Canada (NSERC) Postgraduate Scholarship Doctoral (PGS D) Award and Fonds de recherche du Québec - Nature et Technologies (FRQNT) Doctoral Award. 


\bibliographystyle{unsrtnat}
\bibliography{reference}

\appendix
\newpage
\section{Computational Time} \label{app:time}

\begin{table*}[h]
\caption{Test inference time comparison for snapshot and event based methods on DTDG datasets, we report the average result from 5 runs. Top three models are coloured by \first{First}, \second{Second}, \third{Third}.}
\label{tab:dtdg_inference}
\centering
\resizebox{\linewidth}{!}{%
\begin{tabu}{l | l | l l l l l}
\toprule

 & Method & UCI & Enron & Contacts & Social Evo. & MOOC \\
 \midrule
 \multirow{6}{*}{\rotatebox[origin=c]{90}{\textbf{event}}} & 
 TGN~\cite{tgn2020}                                             & 1.07                                  & 1.71                                  & 137.57            & 24.04                             & 50.04 \\
 & DyGFormer~\cite{dygformer2023}                               & 155.58                                & 57.72                                 & 15423.99          & 349.22                            & OOM \\
 & NAT~\cite{luo2022neighborhood}                               & 3.82                                  & 8.39                                  & 596.22            & 148.43                            & 299.00 \\
 & GraphMixer~\cite{cong2022we}                                 & 32.88                                 & 13.85                                 & 3542.88           & 132.39                            & OOM \\
 & $\text{EdgeBank}_{\infty}$~\cite{poursafaei2022towards}      & 0.52                                  & \first{0.24}                          & \third{45.33}     & \first{2.07}                      & \first{5.17} \\
 & $\text{EdgeBank}_{\text{tw}}$~\cite{poursafaei2022towards}   & 0.52                                  & \second{0.25}                         & 50.77             & \second{2.45}                     & \second{6.12} \\
 \midrule
  \multirow{4}{*}{\rotatebox[origin=c]{90}{\textbf{snapshot}}}  &
 HTGN (\name)~\cite{htgn2021}                                  & 0.61                                  & 0.87                                  & 76.64             & 14.59                             & 28.64 \\  
 & GCLSTM (\name)~\cite{gclstm2022}                             & \first{0.35}                          & 0.46                                  & \second{40.83}    & 9.27                              & 19.78 \\
 & EGCNo (\name)~\cite{evolvegcn2020}                           & \second{0.43}                         & 0.45                                  & \first{40.62}     & 7.35                              & 15.49 \\
 & GCN (\name)~\cite{kipf2016semi}                              & \third{0.50}                          & \third{0.31}                          & 56.88             & \third{6.40}                      & \third{13.30}\\
\bottomrule
\end{tabu}
}
\end{table*}

\begin{table*}[t]
\caption{Test inference time comparison for snapshot and event based methods on CTDG datasets, results reported from 5 runs.  Top three models are coloured by \first{First}, \second{Second}, \third{Third}.}
\label{tab:ctdg_inference}
\centering
\resizebox{0.8\linewidth}{!}{%
\begin{tabu}{l | l | l l l l }
\toprule

 & Method & \texttt{tgbl-wiki} & \texttt{tgbl-review}  & Reddit \\ 
 \midrule

 \multirow{6}{*}{\rotatebox[origin=c]{90}{\small \textbf{event}}} &
 TGN~\cite{tgn2020}                                               & 39.24                   & 1137.69                       & 286.79 \\
 & DyGFormer~\cite{dygformer2023}                                 & 7196.52                 & 26477.51                      & OOT \\
 & NAT~\cite{luo2022neighborhood}                                 & 340.51                  & 8925.21                       & 1159.19 \\
 & GraphMixer~\cite{cong2022we}                                   & 1655.44                 & 4167.63                       & 7166.24 \\
 & $\text{EdgeBank}_{\infty}$~\cite{poursafaei2022towards}        & \second{20.06}          & \first{140.25}                & \first{26.91}\\
& $\text{EdgeBank}_{\text{tw}}$~\cite{poursafaei2022towards}      & 20.67                   & \second{143.49}               & \second{27.08}\\
 \midrule

\multirow{4}{*}{\rotatebox[origin=c]{90}{\small \textbf{snapshot}}} & 
HTGN (\name)~\cite{htgn2021}                                     & 28.96                    & 718.17                        & 117.05 \\  
& GCLSTM (\name)~\cite{gclstm2022}                               & 20.54                    & 436.30                        & 82.88 \\
& EGCNo (\name)~\cite{evolvegcn2020}                             & \third{20.15}            & 433.23                        & 84.85 \\
& GCN (\name)~\cite{kipf2016semi}                                & \first{18.25}            & \third{384.51}                & \third{78.78} \\
 
\bottomrule
\end{tabu}
}
\end{table*}

 Table~\ref{tab:dtdg_inference} shows the inference time for all methods on DTDG datasets. 
 Table~\ref{tab:ctdg_inference} shows the inference time for all methods on CTDG datasets. OOM means out of memory and OOT means out of time. We observe that snapshot-based models are at least one order of magnitude faster than event-based models such as NAT, DyGFormer and GraphMixer. In addition, the best performing model on most datasets, DyGFormer, is also consistently the slowest method.

\section{Evaluation Settings} \label{app:eval}

\begin{figure}[ht]
    \begin{center}
        \centerline{\includegraphics[width=0.9\textwidth]{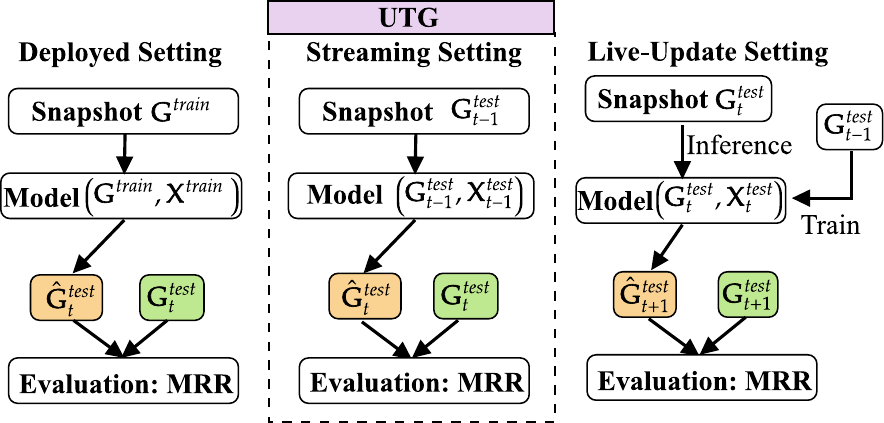}} 
        \caption{Different setting for evaluation of future link prediction include between \emph{deployed}, \emph{streaming} and \emph{live-update} setting. \name framework is designed for the streaming setting.}       
        \label{Fig:eval}
    \end{center}
\end{figure}

\textbf{Deployed setting} The \emph{deployed setting} is often used as the evaluation setting for snapshot-based methods~\cite{htgn2021,evolvegcn2020}. In this setting, no information from the test set is passed to the model, and the node embeddings from the last training snapshot are used for predictions in all test snapshots. 

\textbf{Streaming setting} event-based models often evaluate with the \emph{streaming setting}~\cite{tgn2020,dygformer2023,cawn2021,luo2022neighborhood}. In this setting, information from the previously observed batches of events can be used to update the model however no information from the test set can be used to train the model. 

\textbf{Live-update Setting} \citet{roland2022} proposed the \emph{live-update setting} where the model weights are constantly updated to newly observed snapshots while predicting the next snapshot. To predict links in $\mat{G}_{t+1}$, first the observed snapshot $\mat{G}_{t-1}$ are split into a training set and a validation set. The model is trained on $\mat{G}_{t-1}^{train}$ while using $\mat{G}_{t-1}^{val}$ for early stopping. Lastly, the trained model receives  $\mat{G}_{t}$ and predicts for  $\mat{G}_{t+1}$. 

Figure~\ref{Fig:eval} illustrates the difference between these three settings. In this work, we focus on the streaming setting as it closely resembles the common use case where even after a model is trained, it is expected to incorporate new information from the data stream for accurate predictions.

\section{Temporal Graph Learning Methods} \label{app:method}

\subsection{Snapshot-based Methods}
\changed{snapshot-based methods receives a sequence of graph snapshots as input, representing the temporal graph at specific time intervals (hours, days, etc.). Therefore, DTDG methods are designed to process entire snapshot at once (often with a graph learning model) and then utilize mechanisms to learn temporal dependencies between snapshots. Example methods are as follows:}
\begin{itemize}[leftmargin=*]
    \item \textbf{HTGN.} Many DTDG methods focus on learning structural and temporal dependencies in an Euclidean space thus omitting the complex and hierarchical properties which arises in real world networks. To address this, Yang \etc~\cite{htgn2021} proposed a Hyperbolic Temporal Graph Network~(HTGN) which utilizes the exponential capacity and hierarchical awareness of hyperbolic geometric. More specifically, HTGN incorporates hyperbolic graph neural network and hyperbolic gated recurrent neural network to capture the structural and temporal dependencies of a temporal graph, implicitly preserving hierarchical information. In addition, the hyperbolic temporal contextual self-attention module is used to attend to historical states while the hyberbolic temporal consistency module ensures model stability and generalization. 
    
    \item \textbf{GCLSTM.} To learn over a sequences of graph snapshots, Chen \etc~\cite{gclstm2022} proposed a novel end-to-end ML model named Graph Convolution Network embedded Long Short-Term Memory~(GC-LSTM) for the dynamic link prediction task. In this work, the LSTM act as the main framework to learn temporal dependencies between all snapshots if a temporal graph while GCN is applied on each snapshot to capture the structural dependencies between nodes. Two GCNs are used to learn the hidden state and the cell state for the LSTM and the decoder is a MLP mapping the feature at the current time back to the graph space. The design of GC-LSTM allows it to handle both link additions and link removals.

    \item \textbf{EGCN.} Existing approaches often require the knowledge of a node during the entire time span of a temporal graph while real world networks often changes its node set. To address this challenge, Pareja \etc~\cite{evolvegcn2020} proposed the EvolveGCN~(EGCN) model which captures the dynamic of the graph sequence by using an RNN to update the weight of a GCN. In this way, The RNN regulates the GCN model parameter directly and effectively performing model adaptation. This allows node changes because the learning is performed on the model itself, rather than specific sequence of node embeddings. Note that the GCN parameters are not trained and only computed from the RNN. Empirically, the model achieves good performance for link prediction, edge classification and node classification on DTDGs. 
    
    \item \textbf{PyG-Temporal}. PyTorch Geometric Temporal~(PyG-Temporal) is an open-source Python library which combines state-of-the-art methods for neural spatiotemporal signal processing~\cite{pygtemporal2021}. Many existing methods such as EGCN, GCLSTM and more are implemented directly in PyG-Temporal for research. PyG-Temporal is designed with a simple and consistent API following existing geometric deep learning library such as Pytorch Geometric~\cite{fey2019fast}. Originally, PyG-Temporal are designed for node level regression tasks on datasets available exclusively within the framework. In this work, we apply PyG-Temporal models for the link prediction tasks on publicly available datasets. 
    
\end{itemize}

\subsection{Event-based Methods}

\changed{
Continuous Time Dynamic Graph~(CTDG) methods receive a continuous stream of edges as input and make predictions over any possible timestamps. CTDG methods incorporate newly observed information into its predictions by updating its internal representation of the world. For efficiency, the stream of edges are divided into fixed size batches while predictions are made for each batch sequentially. To incorporate the latest information, edges from each batch becomes available to the model once the predictions are made. Different from DTDG, CTDG has no inherent notion of graph snapshots, models often track internal representations of a node over time and sample temporal neighborhoods surrounding the node of interest for prediction.}
\begin{itemize}[leftmargin=*]
    \item \textbf{TGAT.} Xu \etc~\cite{xu2020inductive} argued that models for temporal graphs should be able to quickly generate embeddings in an inductive fashion when new nodes are encountered. The key component of the proposed Temporal Graph Attention~(TGAT) layer is to combine the self-attention mechanism with a novel functional time encoding technique derived from Bochner's theorem from classical harmonic analysis. In this way, a TGAT layer can efficiently learn from temporal neighborhood features as well as temporal dependencies. The functional time encoding provides a continuous functional mapping from the time domain to a vector space. The hidden vector of time then replaces positional encoding used in the self-attention mechanism. 

    \item \textbf{TGN.} Rossi \etc.~\cite{tgn2020} introduce Temporal Graph Network (TGN), a versatile and efficient framework for dynamic graphs, represented as stream of timestamped events. TGN leverage a combination of memory modules and graph-based operators to improve computational efficiency. Essentially, TGN is a framework that subsumes several previous models as specific instances. When making predictions for a new batch, TGN first update the memory with messages coming from previous batches to allow the model to incorporate novel information from observed batches.
    
    \item \textbf{CAWN.} Causal Anonymous Walks (CAWs)~\cite{cawn2021} are proposed for representing temporal networks inductively to learn the laws governing the link evolution on networks such as the triadic closure law. CAWs, derived from temporal random walks, act as automatic retrievals of temporal network motifs, avoiding the need for their manual selection and counting. An anonymization strategy was also proposed to replace node identities with hitting counts from sampled walks, maintaining inductiveness and motif correlation. 
    CAWN is a neural network model proposed to encode CAWs, paired with a CAW sampling strategy that ensures constant memory and time costs for online training and inference.

    \item \textbf{TCL.} TCL~\cite{wang2021tcl} effectively learns dynamic node representations by capturing both temporal and topological information. It features three main components: a graph-topology-aware transformer adapted from the vanilla Transformer, a two-stream encoder that independently extracts temporal neighborhood representations of interacting nodes and models their interdependencies using a co-attentional transformer, and an optimization strategy inspired by contrastive learning. This strategy maximizes mutual information between predictive representations of future interaction nodes, enhancing robustness to noise.

    \item \textbf{NAT.} In modeling temporal networks, the neighborhood of nodes provides essential structural information for interaction prediction. It is often challenging to extract this information efficiently. Luo \etc~\cite{luo2022neighborhood} propose the Neighborhood-Aware Temporal (NAT) network model that introduces a dictionary-type neighborhood representation for each node. NAT records a down-sampled set of neighboring nodes as keys, enabling fast construction of structural features for joint neighborhoods. A specialized data structure called N-cache is designed to facilitate parallel access and updates on GPUs. 

    \item \textbf{EdgeBank.} EdgeBank~\cite{poursafaei2022towards} is a non-learnable heuristic baseline which simply memorizes previously observed edges. The surprisingly strong performance of EdgeBank in existing evaluation inspired the authors to also propose novel, more challenging and realistic evaluation protocals for dynamic link prediction. 

    \item \textbf{DyGFormer} \textbf{DyGFormer.} Yu \etc~\cite{dygformer2023} introduces a transformer-based architecture for dynamic graph learning. DyGFormer focuses on learning from nodes historical first-hop interactions and employs a neighbor co-occurrence encoding scheme to capture correlations between source and destination nodes through their historical sequences. A patching technique was also proposed to divide each sequence into patches for the transformer, enabling effective utilization of longer histories. \textit{DyGLib} was also presented as a library for standardizing training pipelines, extensible coding interfaces, and thorough evaluation protocols to ensure reproducible dynamic graph learning research.
\end{itemize}


\section{Computing Resources}
\label{sec:computing_resources}

For our experiments, we utilized one of the following GPUs. 
The first option was NVIDIA A100 GPUs (40GB memory) paired with 4 CPU nodes. These nodes featured CPUs such as the AMD Rome 7532 @ 2.40 GHz with 256MB cache L3, AMD Rome 7502 @2.50 GHz with 128MB cache L3, or AMD Milan 7413 @ 2.65 GHz with 128MB cache L3, each equipped with 100GB memory. 
The second option was using NVIDIA V100SXM2 GPUs (16GB memory) alongside 4 CPU nodes, which housed Intel Gold 6148 Skylake CPUs @ 2.4 GHz, each with 100GB memory. 
Our last choice was to run experiments using NVIDIA P100 Pascal GPUs (12GB HBM2 memory) with 4 CPU nodes from Intel E5-2683 v4 Broadwell @ 2.1GHz with 100GB memory.
Each experiment had a five-day time limit and was repeated five times, with results reported as averages and standard deviations. Notably, aside from methods adopted from the PyTorch Geometric library, several other models (assessed using their original source code or the \href{https://github.com/yule-BUAA/DyGLib}{DyGLib repository}) encountered out-of-memory or out-of-time errors when applied to larger datasets.

\section{Model Configurations}
\label{sec:model_config}

For all methods and datasets, we employed the Adam optimizer with a two different learning rates namely \textit{$0.001$} and \textit{$0.0002$}, and the configuration with the higher average performance was selected for reporting the results.
Each experiment was repeated five times and the average and standard deviations were reported. 

The train, validation, and test splits for \texttt{tgbl-wiki} and \texttt{tgbl-review} are provided by the \texttt{TGB} benchmark.
For other datasets (namely, UCI, Enron, Contacts, Social Evo., MOOC, and Reddit), we used a chronological split of the data with \textit{$70\%$}, \textit{$15\%$}, and \textit{$15\%$} for the training, validation, and test set, respectively, which is inline with previous studies \cite{poursafaei2022towards, tgn2020, tgat2020,luo2022neighborhood}. 
We set the batch size equal to 64 for NAT, and for all other models (i.e., TGN, DyGFormer, GraphMixer, EdgeBank, HTGN, GCLSTM, EGCNo, and GCN) the batch size was 200.
For the experiments on CTDGs, we set the number of epoch equal to 40 and implemented an early stopping approach with a patience of 20 epochs and tolerance of \textit{$10^{-5}$}.
For the experiments on DTDGs, the number of epochs was set to 200 with a similar early stopping approach. Dropout was set to 0.1. We set the number of attention heads equal to 2 for the models with an attention module, and node embedding size was fixed at 100.
For TGN, the time embedding size was 100 and the memory dimension was specified as 172, with a message dimension of 100. 
For NAT, we set the \texttt{bias=1e-5}, and \textit{replacement probability=0.7}. All other parameters were set according to the suggested values by \citet{luo2022neighborhood}.
The special hyperparameters of the DyGFormer and GraphMixer are set according to the recommendations presented by \citet{dygformer2023}.

\end{document}